\documentclass[10pt,journal,compsoc]{IEEEtran}
\usepackage{amsfonts}
\usepackage{multicol}
\usepackage{xcolor}

\ifCLASSOPTIONcompsoc
  \usepackage[nocompress]{cite}
\else
  \usepackage{cite}
\fi
\ifCLASSINFOpdf
\else
\fi
\usepackage{amsmath}
\usepackage{hyperref}

\usepackage{algorithmic}

\usepackage{array}
\usepackage{graphicx}
\usepackage{amsthm}

\theoremstyle{definition}
\newtheorem{definition}{Definition}[section]
\usepackage{soul}
\newcommand{\review}[1]{#1}

\begin{document}
%
\title{Transformer-based Graph Neural Networks for Outfit Generation}
%
%
%
%

\author{Federico Becattini,
        Federico Maria Teotini,
        and Alberto Del Bimbo
\IEEEcompsocitemizethanks{\IEEEcompsocthanksitem F. Becattini, F. M. Teotini and A. Del Bimbo are with the Media Integration and Communication Center (MICC) of the University of Florence, Italy.\protect\\
E-mail: name.surname@unifi.it
}
}

%
%

\markboth{Journal of \LaTeX\ Class Files,~Vol.~14, No.~8, August~2015}%
{Shell \MakeLowercase{\textit{et al.}}: Bare Demo of IEEEtran.cls for Computer Society Journals}
%



\IEEEtitleabstractindextext{%
\begin{abstract}
Suggesting complementary clothing items to compose an outfit is a process of emerging interest, yet it involves a fine understanding of fashion trends and visual aesthetics. Previous works have mainly focused on recommendation by scoring visual appeal and representing garments as ordered sequences or as collections of pairwise-compatible items. This limits the full usage of relations among clothes. We attempt to bridge the gap between outfit recommendation and generation by leveraging a graph-based representation of items in a collection. The work carried out in this paper, tries to build a bridge between outfit recommendation and generation, by discovering new appealing outfits starting from a collection of pre-existing ones. We propose a transformer-based architecture, named TGNN, which exploits multi-headed self attention to capture relations between clothing items in a graph as a message passing step in Convolutional Graph Neural Networks. Specifically, starting from a seed, i.e.~one or more garments, outfit generation is performed by iteratively choosing the garment that is most compatible with the previously chosen ones. Extensive experimentations are conducted with two different datasets, demonstrating the capability of the model to perform seeded outfit generation as well as obtaining state of the art results on compatibility estimation tasks.
\end{abstract}

}

\maketitle

\IEEEdisplaynontitleabstractindextext

%
\IEEEpeerreviewmaketitle

\IEEEraisesectionheading{\section{Introduction}}
Fashion is one of the most important industries in the world, moving very large amounts of money with thousands of customers and field operators. Choosing garments to create a good looking outfit is a very important task in the fashion field, but it is a difficult one: it involves many complex concepts like style and visual composition expertise, creativity, cultural and social understanding, trends, etc., and they all need to be balanced to make sure that the resulting outfit is indeed aesthetically appealing.

Outfit creation is an ubiquitous task and, with the rapid growth of online fashion retailers and fashion related social networks (e.g.~Instagram), it became a fundamental task, often carried out by entire company departments of fashion experts.

Every architecture needs to support two notions in order to put together a good outfit:

\begin{itemize}
\item
  The \emph{similarity} notion, that is when two garments are similar to each other and possibly interchangeable.
\item
  The \emph{compatibility} notion, meaning that the fashion items composing the same outfit should aesthetically be compatible with each other.
\end{itemize}

\begin{figure}[t]
\centering
\includegraphics[width=0.8\columnwidth]{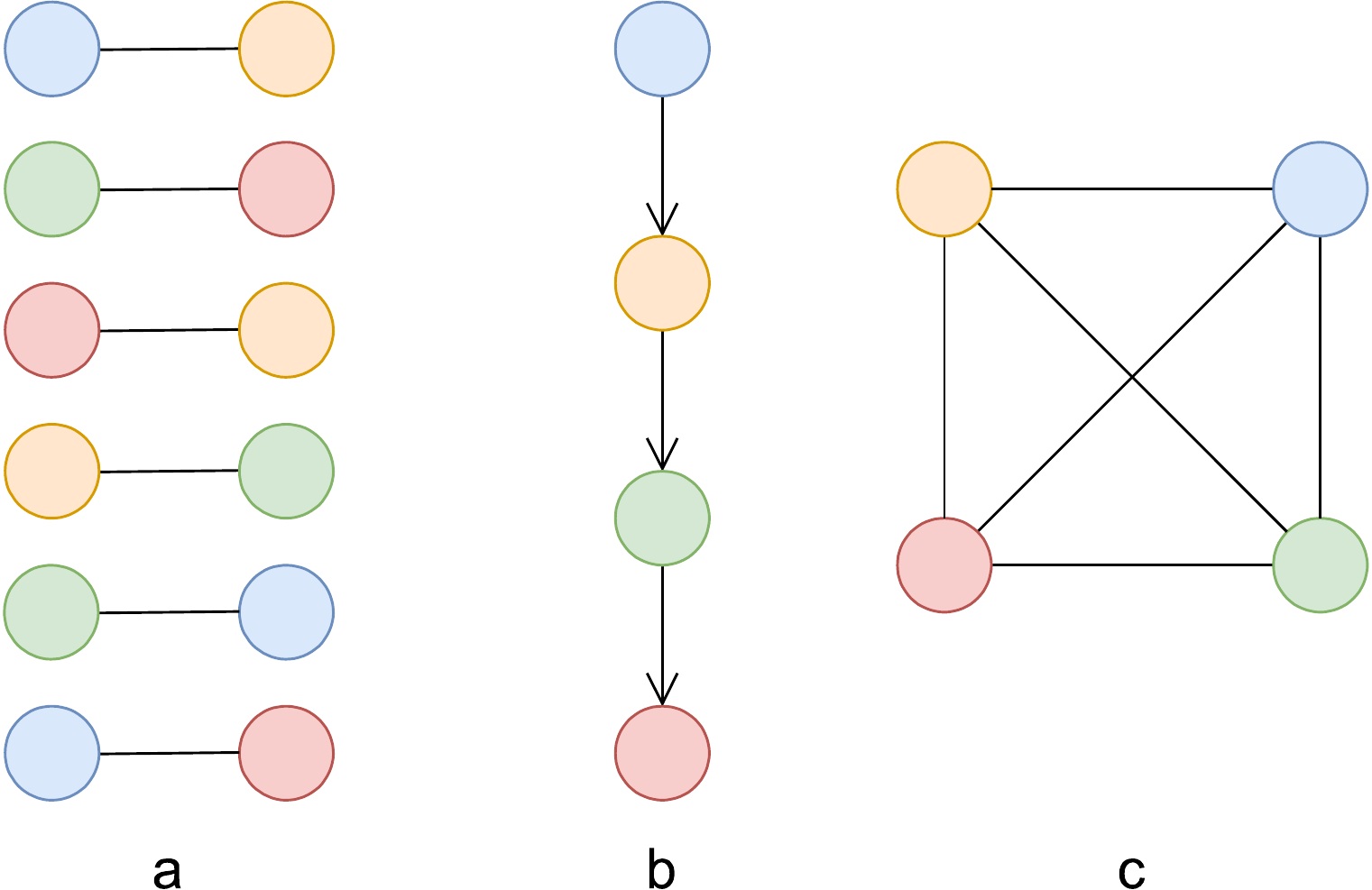}
\caption{Different representations for an outfit composed by four different items: (\textbf{a}) the pairwise one, (\textbf{b}) the ordered sequence one, and (\textbf{c}) the graph one.}
\label{fig:out-repr}
\end{figure}

Previous studies \cite{mcauley2015imagebased}, \cite{vasileva2018learning} focused on learning compatibility metrics between pairwise items (Figure \ref{fig:out-repr} \textbf{a}); these kind of architectures cannot model the complex relations between outfit items since each pair is treated independently. Some works \cite{han2017learning} attempted to represent an outfit as an ordered sequence of garments (Figure \ref{fig:out-repr} \textbf{b}) and using Recurrent Neural Network (RNN) to model compatibility. This kind of representation, however, is a forcing as it is reasonable to consider an outfit as a set, thus without a specified order, instead of as a list. To overcome these representation limitations, further architectures \cite{cui2019dressing}--\cite{cucurull2019contextaware} explored the concept of representing an outfit as a graph (Figure \ref{fig:out-repr} \textbf{c}).

In this paper a new architecture is proposed, \emph{Transformer-based Graph Neural Network} (TGNN), aimed at generating new outfits starting from a garment or a set of garments and thus allowing even inexperienced people to create their own outfit. TGNN is based on two complementary architectures: the Transformer \cite{vaswani2017attention} and Graph Neural Network \cite{gori2005new}, \cite{scarselli2009graph}. Transformer is a kind of encoder-decoder architecture \cite{cho2014learning}, \cite{sutskever2014sequence} born in the Natural Language Processing (NLP) field and it is the base architecture for many state-of-art NLP models. GNN proved successful in modeling complex relations in very large graphs \cite{hamilton2017inductive}, \cite{kipf2017semisupervised} and is one of the most researched fields at the time of writing.\\

 \begin{figure}[t]
 {\centering \includegraphics[width=0.7\linewidth]{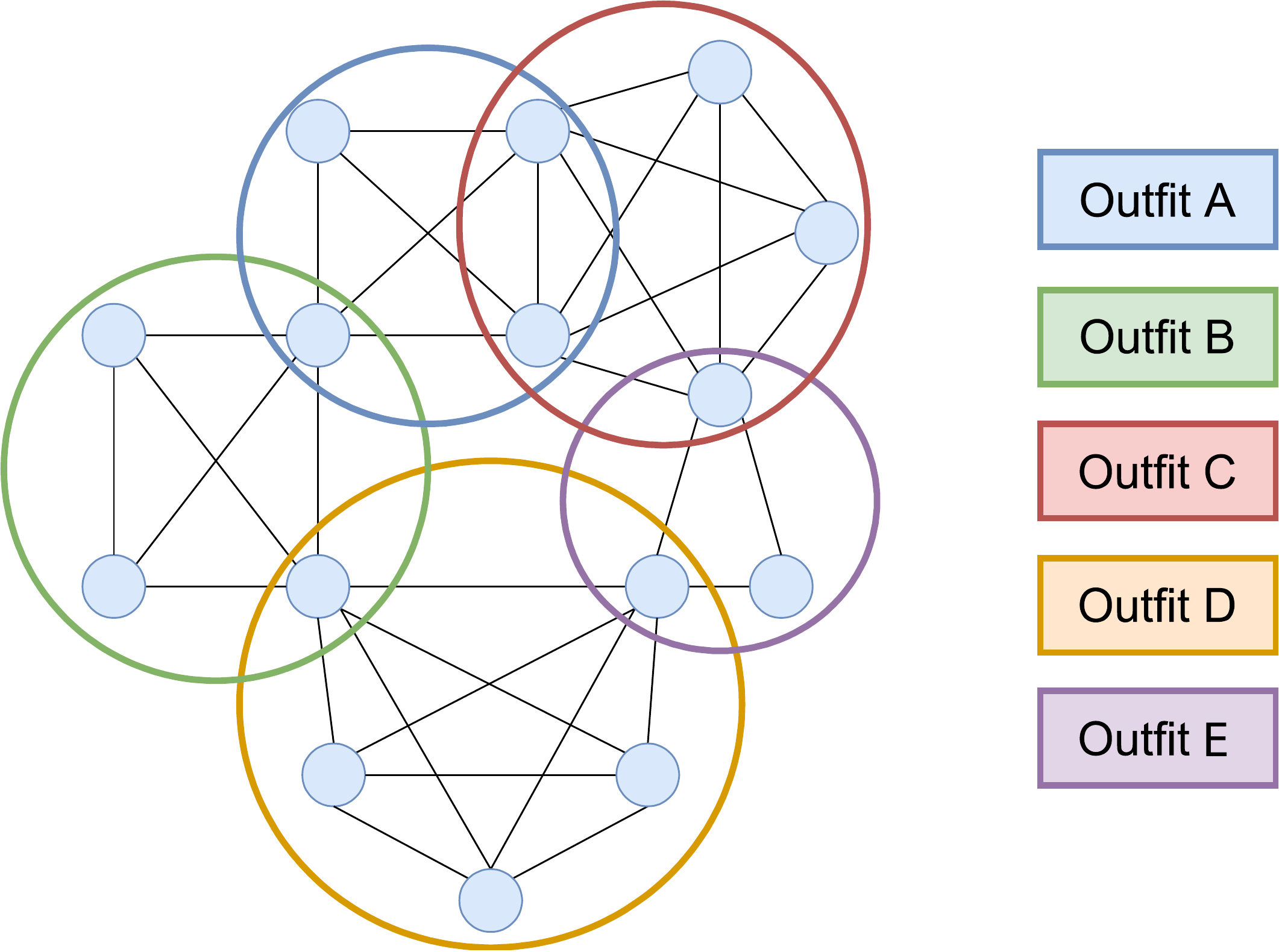} 
 }
 \caption{The item relation graph. There are five different highlighted outfits, each connected to another at least through one or more shared items. In particular, each item is linked to all the other ones appearing in the same outfits.}\label{fig:item-graph}
 \end{figure}

The GNN part of TGNN is aimed at learning context features from this graph. The Transformer architecture, on the other hand, is employed for its outstanding performance on sequences modeling. Born to overcome the recurrent nature of architectures like LSTM \cite{hochreiter1997long} and GRU \cite{cho2014learning}, the Transformer is capable of predicting the next item in a sequence, given the previous ones; this ability is a perfect fit for the outfit generation task starting from a seed: every predicted garment has to be compatible with the seed and the already predicted garments.\\
It has been proved that, under some constraints, GNNs can be seen as Transformer encoders \cite{chaitanya2020transformers}. TGNN uses this concept to integrate these two architectures and to successfully model the compatibility in fashion outfit generation.

To summarize, the main contributions of this work are:

\begin{itemize}
\item
  The proposal of Transformer-based Graph Neural Network (TGNN), a new architecture aimed at seeded outfit generation, which can better capture the complex relations among multiple items in an outfit.
\item
  The development of an hybrid between Graph Neural Networks and Transformer encoders, treating the nodes' neighborhoods as interlinked unordered sequences.
\item
  Experimental results on widely used evaluation tasks demonstrate the effectiveness of the developed architecture over other previous state-of-the-art ones.
\item
  The introduction of a new evaluation task, Seeded Items Prediction (SIP), to test how good a model is at iteratively reconstructing a given outfit starting from one outfit item (the seed).
\end{itemize}

\section{Related work}\label{related}



Fashion recommendation is a hot topic, for which several aspects have been studied. The problem has been declined in many forms, ranging from compatibility estimation~\cite{song2017neurostylist, song2019gp, divitiis2021garment, de2022disentangling, becattini2022fashion}, outfit generation~\cite{chen2019pog} to plain simple recommendation~\cite{lin2019explainable, de2021style} and try-on~\cite{morelli2022dress, zheng2019virtually}.

Many early works \cite{mcauley2015imagebased}, \cite{liu2016deepfashion}--\cite{lu2019learning} focus on calculating a pairwise compatibility metric between garments: for example, \emph{McAuley et al.} \cite{mcauley2015imagebased} extracts visual features to model human visual preference for a pair of items, \emph{Veit et al.} \cite{veit2015learning} develops a Siamese based network that estimates pairwise compatibility based on co-occurrence in large-scale user behavior data, while \emph{Lu et al} \cite{lu2019learning} aggregates user preferences on each item and integrates them with pairwise compatibility scores. These approaches lack efficiency and accuracy in real-world usage.

The first work considering a fashion compatibility between a set of items is \emph{Han et al.} \cite{han2017learning}: here an outfit is considered as an ordered sequence and fed to a bidirectional LSTM network to predict a compatibility score. In this work the Polyvore dataset was also introduced. \emph{Vasileva et al.} \cite{vasileva2018learning} trains pairwise embedding spaces to learn different representations for different pairs of categories, that, however, is not feasible when the number of categories is high. Furthermore, they also extended and enriched the Polyvore dataset introducing more challenging evaluation sets.

Some works \cite{cui2019dressing}, \cite{cucurull2019contextaware}, \cite{yang2020learning} focused on representing an outfit as a graph: in NGNN \cite{cui2019dressing}, \emph{Cui et al.} represented an outfit as a graph to model complex relations among items which has been demonstrated to be more effective than pairwise and sequence representations.
In our work we represent outfits as graphs, treating fashion items as nodes, connected to each other if they can be combined to compose an outfit. Differently from prior work, we process such graphs with a model harnessing the effectiveness of Graph Convolutions as well as multi-headed self-attention, typical of transformers~\cite{vaswani2017attention}. To this end, we integrate the message passing strategy of Graph Convolutions as a form of attention in the encoder structure of a transformer model.
Transformers have also been used recently in literature to generate outfit level representations for compatibility prediction tasks~\cite{sarkar2022outfittransformer}.

In addition to visual features, some take advantage of other kind of information like textual representations \cite{vasileva2018learning}, \cite{cui2019dressing}, \cite{chen2019pog}, categories \cite{vasileva2018learning}, \cite{li2020hierarchical}, and user preferences \cite{li2020hierarchical}, \cite{lu2019learning}, \cite{becattini2021plm}. Lately, there have been studies \cite{revanur2021semisupervised} aimed at operating in semi-supervised settings, as creating fully labeled datasets is expensive and often requires fashion experts' knowledge.

\section{Overview}\label{overview}

Transformer-based Graph Neural Network (TGNN) is a novel architecture whose objective is to discover new aesthetically appealing outfits by picking and combining items from a collection.
When referring to an outfit we refer to a composition of several garments or accessories of different categories that can be worn together. For instance, an outfit can be composed by combining a t-shirt, a skirt, a scarf, shoes, a bag and a necklace.
Starting from an initial garment seed, TGNN is capable of generating unseen outfits complementing the given item. TGNN will iteratively chose garments from a candidate set that are compatible with the seed and with the other previously chosen garments, until a stop condition is reached.
\review{
We make the assumption that the collection of garments from which to make a recommendation is fixed.
}

The TGNN architecture is based on a Transformer network: the encoder block operates on a collection of pre-defined outfits to learn the complex relations standing between garments, while the decoder carries out the outfit generation, conditioning the output with the seed and the contextual information provided by the encoder.

Following prior work, we represent an outfit as a graph clique. Since a single garment can belong to more than one outfit, the collection of pre-defined outfits is naturally modeled as a graph, connecting such cliques through nodes belonging to more outfits.
The encoder blocks in TGNN are therefore a special kind of Graph Neural Network that adapts the standard transformer encoder to graph data structures.

\subsection{Notation and Problem Definition}\label{notation-and-definitions}

Let $\mathcal{O}$ be a collection of pre-defined outfits. Each outfit $o_i \in \mathcal{O}$ is composed of a combination of $n_i$ garments $g_{i,j}$ with $j=1,...,n_i$ belonging to a set $\mathcal{G}$, i.e. $o_i=\{g_{i,1},\cdots,g_{i,n_i}\}$. Each item belongs to a distinct fashion category, which we identify as $C(g_{i,j})$. Since the same garment can belong to several different outfits, we denote with $O(g_i) := \{o_j\ \vert\ g_i \in o_j \}$ the set of outfits in which the item appears. Every outfit might be composed of a different number of garments, depending on how many fashion categories have been used.

In order to build the data structure representing the collection of outfits, we provide the following definitions.
\begin{definition}[Outfit Relation Graph]
\label{def:org}An Outfit Relation Graph (ORG) is an undirected graph whose nodes are outfits and edges exist between two nodes if the linked outfits share at least one garment. Formally, $G_{\mathcal{O}}=(V,\mathcal{E})$ such that
\begin{gather*}
    V \subseteq \mathcal{O} \\[5pt]
    \mathcal{E} =\{(o_i,o_j)\ |\ o_i \cap o_j \ne \emptyset\, ,\, i \ne j\}
\end{gather*}
\end{definition}

\begin{definition}[Item Relation Graph]
\label{def:irg}An Item Relation Graph (IRG) is an undirected graph whose nodes are garments and edges exist between two nodes if the linked garments belong to the same outfit. Formally, \(G_{\mathcal{G}}=(V,\mathcal{E})\) such that
\begin{gather*}
    V \subseteq \mathcal{G} \\[5pt]
    \mathcal{E} =\{e_{i,j}:=(g_i,g_j)\ |\ O(g_i) \cap O(g_j) \ne \emptyset\}
\end{gather*}
When dealing with an IRG, the direct neighborhood $\mathcal{N}_i=\{g_j\,|\, g_j \in V,\, e_{i,j} \in \mathcal{E}\}$ of node $g_i$ is assumed to contain $g_i$ itself, that is, an IRG is an undirected graph with self edges. An example of Item Relation Graph is shown in Fig. \ref{fig:item-graph}.
\end{definition}

\begin{definition}[ORG to IRG induction]
\label{def:org-to-irg}An ORG can induce an IRG, meaning, an IRG can be built starting from an ORG. Formally, given an ORG \(G_{\mathcal{O}}=(V,\mathcal{E})\), its induction \(\overline{G_{\mathcal{O}}}=(\overline{V},\overline{\mathcal{E}})\) is an IRG such that
\begin{equation*}
\begin{split}
    \overline{V}&=\{g_i\ \vert\ O(g_i) \cap V \ne \emptyset \} \\[5pt]
    \overline{\mathcal{E}}&=\{e_{i,j}:=(g_i,g_j)\ |\  g_i,g_j \in \overline{V}, O(g_i) \cap O(g_j) \ne \emptyset\}
\end{split}
\end{equation*}
\end{definition}

Based on these concepts, we define the problem of Seeded Item Prediction as the task aimed at generating complementary items to compose an outfit, based on a garment seed and on an IRG $G_{\mathcal{G}}$. 
A garment seed is a sequence of garments appearing in $G_{\mathcal{G}}$, such that they do not belong to the same outfit, i.e., they are not linked. Formally
\begin{equation*}
    \varphi=\{g_{s,1},\cdots,g_{s,n}\ |\ g_i \in V\, , e_{i,j} \notin \mathcal{E}\quad \forall\, i \ne j\}
\end{equation*}



\section{Methodology}\label{methodology}

Let $G_{\mathcal{G}}$ be an IRG, $\varphi$ a garment seed, and $\mathcal{C} \subset \mathcal{G}$ the garment candidate set with $\varphi \cap \mathcal{C} = \emptyset$ and $\overline{g_\omega} \in \mathcal{C}$, where $\overline{g_\omega}$ is a fake garment indicating end-of-outfit, to be used as a stop sign during generation.

At each time step $t$, TGNN predicts the next outfit garment $\widehat{g}_t$, given $G_{\mathcal{G}}$, $\varphi$ and all the previous generated garments $\{\widehat{g}_i\}_{i=1}^{t-1}$. The element $\widehat{g}_t$ is chosen among the elements of 
$\mathcal{C}^{(t)} = \mathcal{C} / \{g \in O(\widehat{g}_i)\}_{i=1}^{t-1}$.
The outfit generated by TGNN at the end of step $t$ is:
\begin{equation}
    \widehat{o}^{(t)}=\varphi\cup\{\widehat{g}_i\}_{i=1}^{t}\qquad \widehat{g}_i \in \mathcal{C}\, ,\widehat{g}_i\ne\widehat{g}_j
    \label{eq:gtn-gen-outfit}
\end{equation}
This process continues until the model outputs the stop sign $\overline{g_\omega}$. The final outfit is therefore $\widehat{o}^{(T-1)}$, where $T$ is the total number of generation steps.

 \begin{figure}[t]
 {\centering \includegraphics[width=0.9\linewidth]{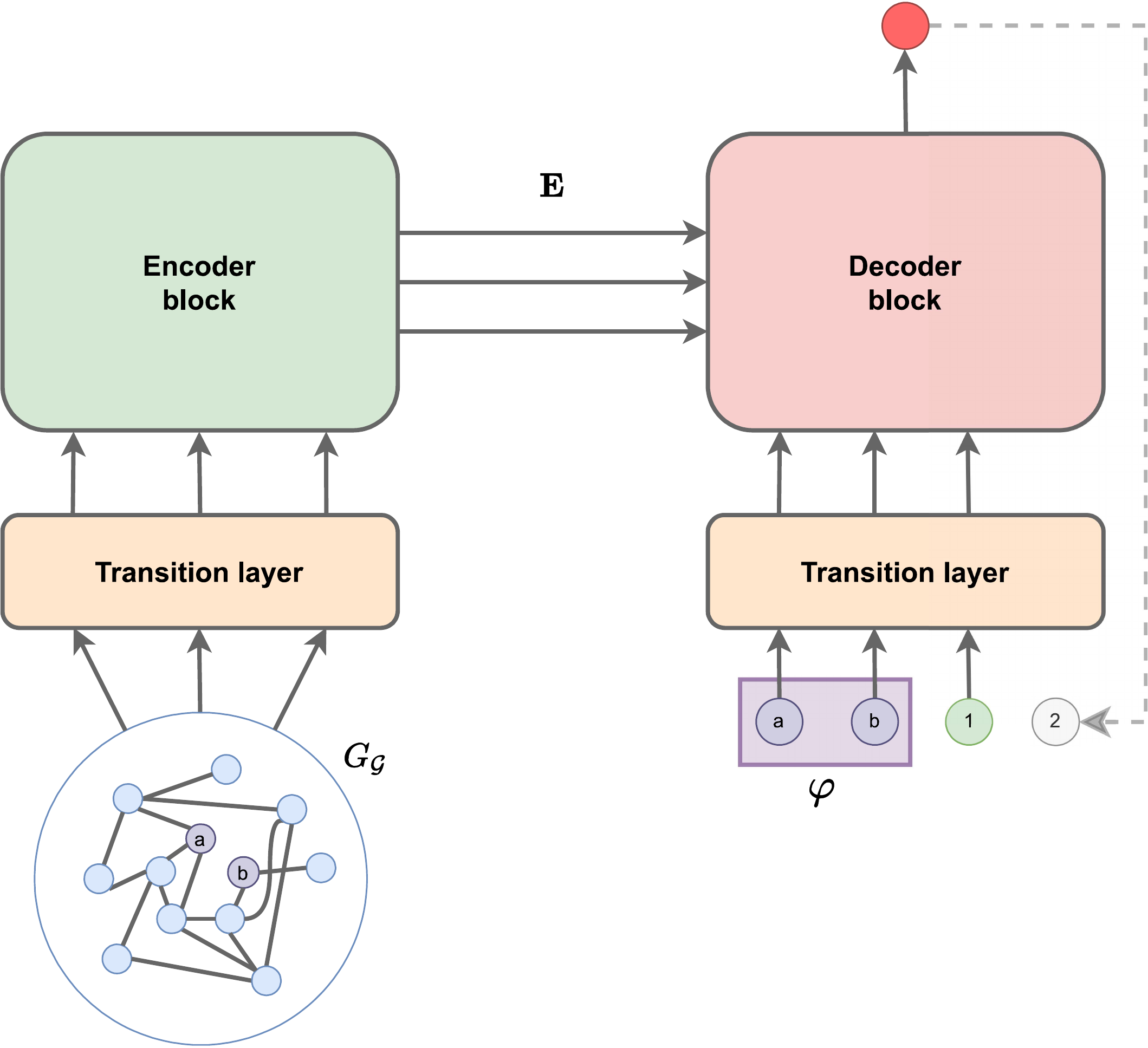} 
 }
 \caption{The architecture of TGNN, here visualized during computation at time step \(t=2\) with a garment seed \(\varphi\) of size \(2\). The garment seed with the previous generated garment \(\widehat{g}_1\) and \(\mathbf{E}\) are fed into the decoder part. The results will become the next outfit garment \(\widehat{g}_2\).}\label{fig:gtn-arch}
\end{figure}

The full TGNN architecture is shown in Fig. \ref{fig:gtn-arch}. The node embeddings of $G_{\mathcal{G}}$ are first passed to a transition layer whose responsibility mirrors the embedding layer of a standard Transformer. Moreover it adapts the $d_e$-dimensional embeddings to the hidden dimension $d_m$ of the model.
Then, the encoder block operates on this modified graph to generate a new representation $\mathbf{E}$ of all the nodes/garments.

For the decoder part, at first, $\widehat{o}^{(t)}$ is passed through another transition layer, then it is fed to the decoder block along with $\mathbf{E}$. The result is a vector $\mathbf{h}^{(t)} \in \mathbb{R}^{d_m}$. Then $\widehat{g}_t$ is obtained as:
\begin{equation}
    \widehat{g}_t = \arg\max_{g_c \in \mathcal{C}^{(t)}}\; \frac{\exp(\mathbf{h}^{(t)} \cdot \tau(\mathbf{g}_c))}{\sum_{g_i \in \mathcal{C}^{(t)}} \exp(\mathbf{h}^{(t)} \cdot \tau(\mathbf{g}_i))}
    \label{eq:gtn-next-garment}
\end{equation}
that is, the garment \(g_c\) of the candidate set, whose transition representation \(\tau(\mathbf{g}_c)\) results in the highest similarity with \(\mathbf{h}^{(t)}\), among all the candidate set items.

Note that in principle the number of candidates could become arbitrarily large. However, each element is independently fed to the transition layer and compared to the output of the decoder, as detailed in Eq. \ref{eq:gtn-next-garment}. This operation can be done easily for large candidate sets, since both operations can be carried out in parallel. If the number of candidate items is so high to make parallel computation unfeasible, an index (e.g. FAISS \cite{johnson2019billion}) could be used to approximate the similarity computation.

In addition, in this paper we assume to have a closed collection of garments from which to make recommendations. Adding new garments would require to update the graph, which could be done at any time without requiring any further training.

\subsection{Initial Garment Embeddings}\label{garm-embeddings}

Unlike Transformers for NLP~\cite{vaswani2017attention}, which deal with words, it is not possible to build a garment ``vocabulary'' containing all the existing garments. On the other hand, a garment can be associated with an image and/or a title. This means that instead of learning the initial embeddings as it is done with words, in the case of garments, they can be pre-computed as feature vectors.

In this work, each garment image is fed into an ImageNet pretrained ResNet50, and the outputs of the last layer before the classification block are then taken as garment embeddings. Since these embeddings are computed by ResNet50, their dimension is $2048$. To compact the features we use PCA and project them to a lower dimension $d_e$.


\subsection{Transition Layer}\label{gtn-trans-layer}

As shown in Fig. \ref{fig:gtn-arch}, there are two transition layers, one for the encoder block and one for the decoder block. These are simply feed-forward layers followed by a ReLU activation, that serve the purpose to learn a meaningful and compact representation to be fed to each block of the model. In particular, for the decoder, it also allows to map items in the candidate set into a semantic space suitable for identifying complementary items according to Eq. \ref{eq:gtn-next-garment}.


\begin{figure}[t]
 {\centering \includegraphics[width=0.9\linewidth]{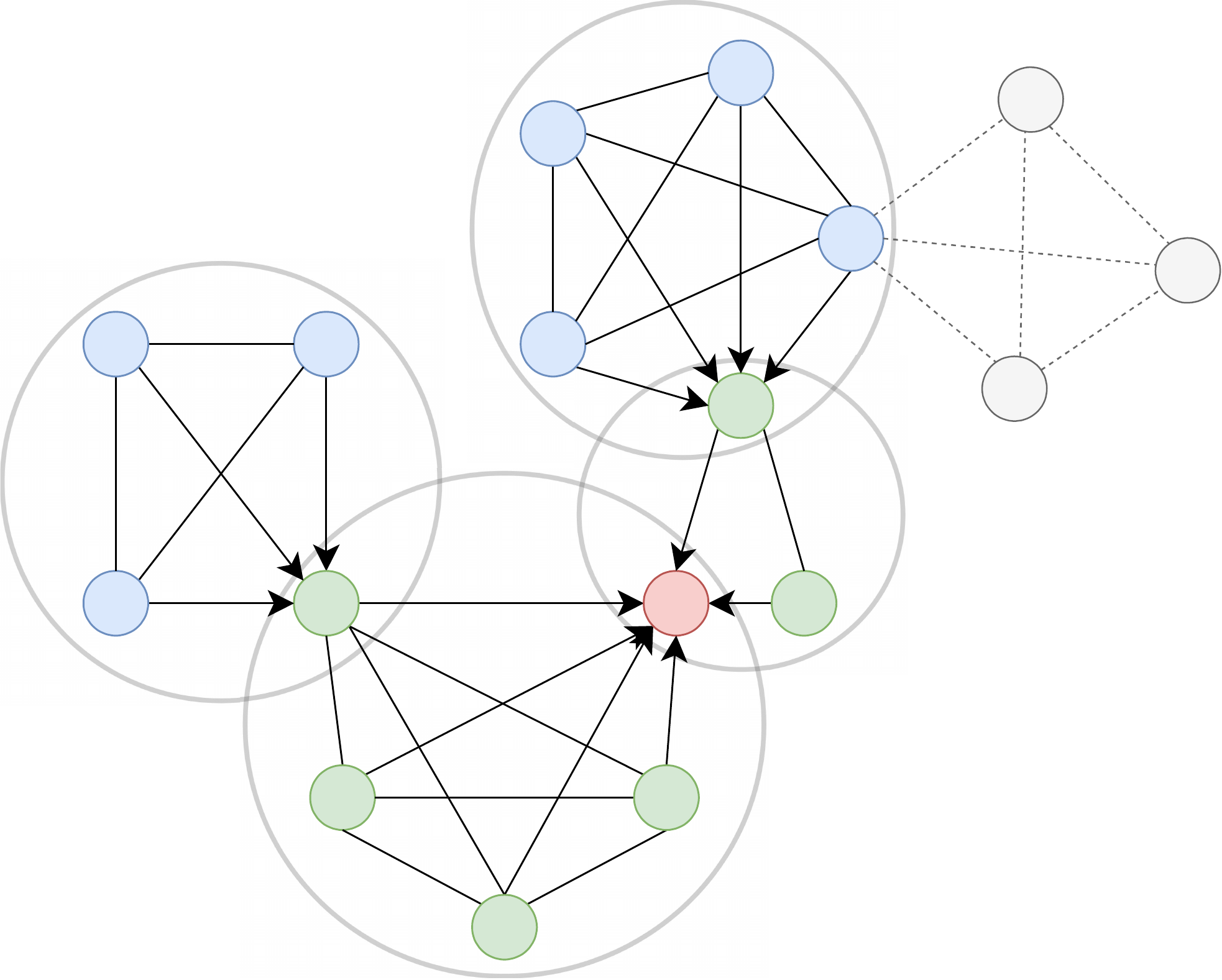} 
 }
 \caption{An example illustration of how attention flows towards the red node in a $K_{enc}=2$ encoder block. The information first flows from the green nodes to the red one, then from the blue nodes, through the green ones, up to the red one. Since $K_{enc}=2$, no attention information coming from the grey nodes can arrive to the red one.}\label{fig:irg-msg}
 \end{figure}

\subsection{Transformer Encoder as ConvGNN}\label{transformer-encoder-as-convgnn}

The encoder block operates on a graph containing a pre-defined collection of outfits, specifically an Item Relation Graph. The traditional encoder of a transformer is modified only in its self-attention layer: the idea, adapted from GAT~\cite{velickovic2018graph}, is to treat each node along with its direct neighbors as an unordered sequence and to apply to each of them the multi-head scaled dot-product attention mechanism. These sequences are actually outfits and, as outfits can share garments, these sequences are linked together. By stacking multiple of these modified encoder modules, the message passing nature of spatial ConvGNNs can be leveraged, meaning that a single garment can attend not only to the garments with which it shares a common outfit, but also to other garments belonging to linked outfits.

In Fig.~\ref{fig:irg-msg}, an example of how attention information is passed between different but linked outfits in an IRG, is shown.



Formally, referring to the MultiHead attention function adopted in transformers \cite{vaswani2017attention}, the multi-head scaled dot-product attention mechanism calculated for a node $\mathbf{g}_i \in \mathbb{R}^d$ is defined as:
\begin{equation}
    \text{MultiHead}(\mathbf{g}_i, \mathcal{N}_i, \mathbf{g}_i)
    \label{eq:gtn-enc-conv}
\end{equation}
that is, the query $Q$ and the value $V$ are set to be equal to $\mathbf{g}_i$ while the key $K$ is a $\vert \mathcal{N}_i \vert \times d$ matrix containing the feature vectors of the neighboring nodes of $\mathbf{g}_i$.

\subsection{Objective Function}\label{loss}

Let $o=\{g_1,\cdots,g_n\}$ be a predefined outfit and let $G_{\mathcal{G}}=(V,\mathcal{E})$ be an IRG where $g_1,\cdots,g_n \in V$ but $e_{i,j} \notin \mathcal{E}\; \forall\, g_i,g_j \in o$, and $\mathcal{C} \subset \mathcal{G}$ the garments candidate set with $o \subset \mathcal{C}$. Thus, for the triplet $(o, G_{\mathcal{G}}, \mathcal{C})$ the objective function of TGNN can be written as
\begin{equation}
    \mathcal{L}_{(o,G_{\mathcal{G}},\mathcal{C})}=-\frac{1}{n}\sum_{t=1}^n \log P(g_{t+1} \vert g_1,\cdots,g_t,G_{\mathcal{G}},\mathcal{C},\Theta)
    \label{eq:gtn-obj}
\end{equation}
where $\Theta$ denotes the model parameters, $g_{n+1}$ is assumed to be the end-of-outfit fake garment $\overline{g_\omega}$, and $P( \cdot )$ is the probability of choosing the next correct garment conditioned on the previously predicted outfit garments, the IRG and the candidate set.

\begin{figure}[t]
 \centering \includegraphics[width=0.8\linewidth]{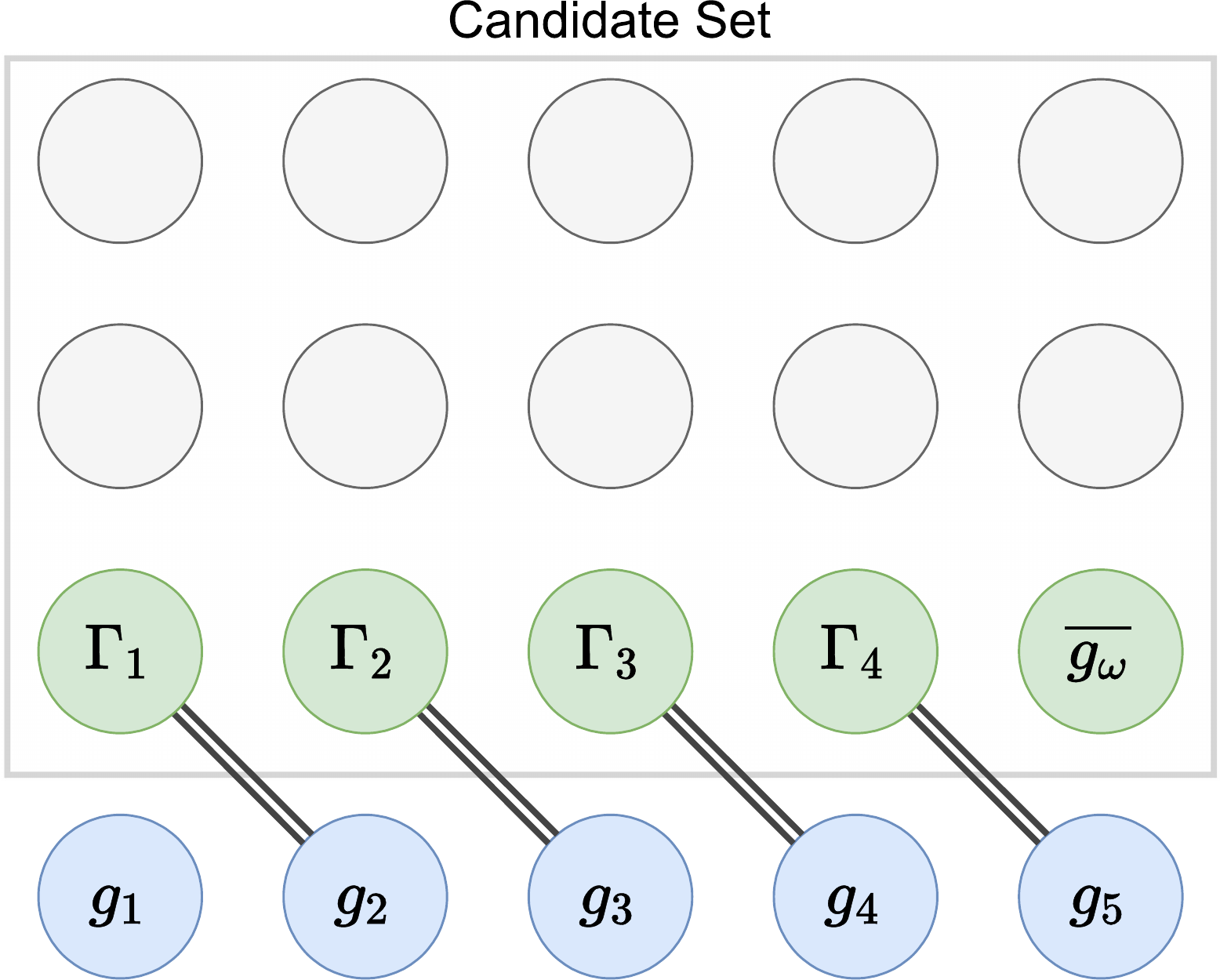}
 \caption{Example illustration of a training sample outfit with five garments (in blue) and their corresponding ground truth items belonging to the candidate set. The ground truth for a given item is equal to the next one, while for the last is the end-of-outfit fake garment.}\label{fig:gtn-obj}
 \end{figure}

In one training sample, the i-th ground truth $\Gamma_i$ is the next garment to be produced, that is $\Gamma_i = g_{i+1}$, as shown in Fig. \ref{fig:gtn-obj}, and \(\Gamma_n = \overline{g_\omega}\).

In this setting, the conditioned probability of Equation \eqref{eq:gtn-obj} becomes
\begin{equation}
    P(g_{t+1} \vert g_1,\cdots,g_t,G_{\mathcal{G}},\mathcal{C},\Theta) = \frac{\exp(\mathbf{h}^{(t)} \cdot \tau(\Gamma_t))}{\sum_{g_i \in \mathcal{C}} \exp(\mathbf{h}^{(t)} \cdot \tau(\mathbf{g}_i))}
    \label{eq:gtn-cond-prob}
\end{equation}

\section{Experiments}\label{experiments}

In this section we outline the experimental validation of the proposed method and the analysis of the obtained results. In addition, both the network setup and the hyperparameter values used throughout the experiments are explained.

\section{Dataset}\label{polyvore}

To perform experimental validation of our approach, we adopt different versions of the Polyvore dataset \cite{vasileva2018learning}.

The dataset has been built from the Polyvore website, a popular social network hub where fashion enthusiasts could upload, tag and title garment images and create custom outfits as compositions of these. The crafted outfits, then, could be shared, commented, liked, but also used as basis for other new outfits.

\begin{figure}[t]
{\centering \includegraphics[width=\linewidth]{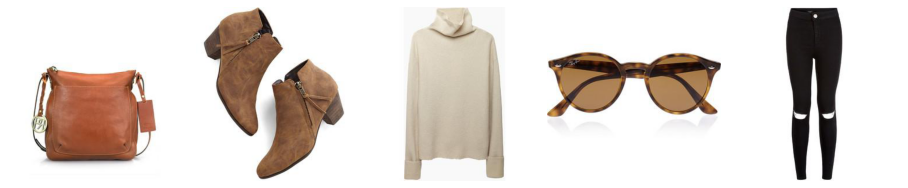} 
}
\caption{An outfit example from the Polyvore dataset \cite{vasileva2018learning}}\label{fig:polyvore-outfit-example}
\end{figure}

An eary version of the dataset was proposed by Han et al. \cite{han2017learning}, which is now referred to as the Maryland Polyvore dataset. This dataset was relatively small (around 20k outfits and 160k items), did not contain item categories, and had some inconsistencies in the test set.
To resolve these issues, Vasileva et al. \cite{vasileva2018learning} produced another outfit dataset based on Polyvore data, referred to as the Polyvore dataset, with both coarse and finer grained item categories (defined respectively as ``semantic'' and ``leaf'' categories), titles and descriptions; this dataset is, also, larger than the Maryland dataset as it contains more than 60K outfits and 360K items. Moreover, this dataset provides carefully tailored test-train splits: as some garments appear in many different outfits, the choice of letting these garments appear in unseen test outfits has a significant effect.
The Polyvore Dataset comes in two versions:

\begin{itemize}
\item
  \textbf{Polyvore-S} (Standard) -- This version is easier as no outfit appearing in the training set or in the test set can appear in the other one, while it is possible that an item belonging to a training outfit is seen in a test one.
\item
  \textbf{Polyvore-D} (Disjoint) -- A more challenging version, where even garments are not allowed to appear in more than one set.
\end{itemize}

\begin{figure}[t]
 {\centering \includegraphics[width=0.95\linewidth]{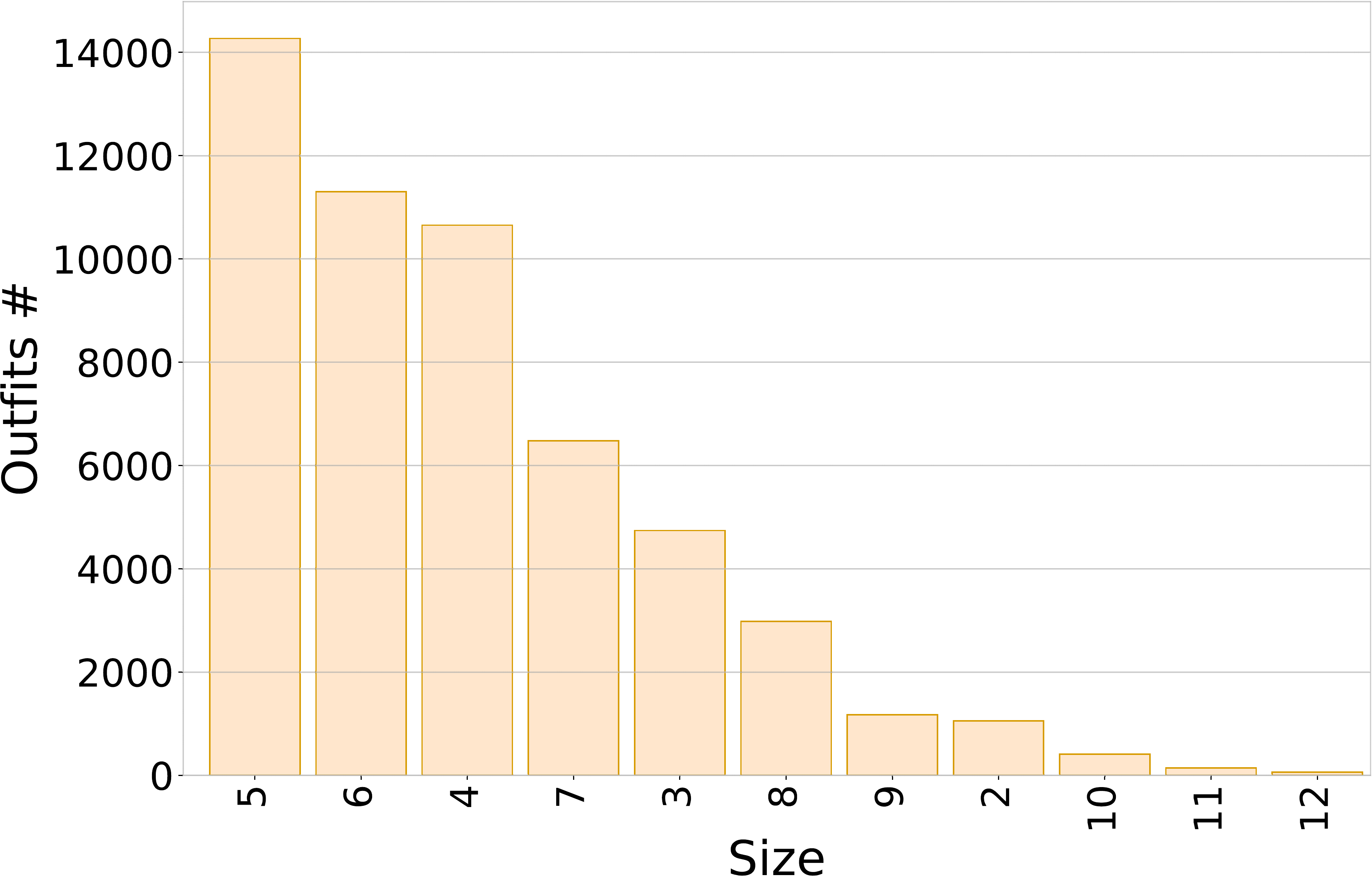} 
 }
 \caption{The outfit size distribution of Polyvore-S train split}\label{fig:polyvore-size-distrib}
 \end{figure}

In Tab. \ref{tab:polyvore-versions}, the outfit and item numbers, along with the split sizes, are summarized for both versions. In Fig. \ref{fig:polyvore-size-distrib}, the outfit size distribution for the Polyvore-S train split, is visualized.

\begin{table}[t]
\caption{\label{tab:polyvore-versions}Size comparison of Polyvore dataset versions}
\centering
\resizebox{\linewidth}{!}{
\begin{tabular}[t]{lccccc}
\hline
Version & \#Outfits & \#Items & Train Outfits & Val Outfits & Test Outfits\\
\hline
\textbf{Polyvore-S} & 68306 & 365054 & 53306 & 5000 & 10000\\
\textbf{Polyvore-D} & 32140 & 175485 & 16995 & 3000 & 15145\\
\end{tabular}
}
\end{table}

\section{TGNN setup}\label{gtn-setup}
First, each garment image is distilled in a feature vector, as described in Section \ref{garm-embeddings}, by feeding it to a pretrained ResNet50 \cite{he2016deep}, and then through a PCA resulting in $d_e=128$ dimensional feature vectors.

Let $G_{\mathcal{G}}$ be the Item Relation Graph of the dataset, as in Definition \ref{def:irg}. Generally speaking, every dataset of interest will have an IRG too large to handle. In Tab. \ref{tab:polyvore-irg}, some graph statistics for the Polyvore IRGs are shown. For example, the Polyvore-S train split IRG contains more than 200k nodes and more than 680k edges.

\begin{table}[t]
\caption{\label{tab:polyvore-irg}Graph statistics for both the train and test IRGs of the Polyvore-S and Polyvore-D datasets}
\centering
\begin{tabular}[t]{lrrrr}
\hline
\multicolumn{1}{c}{ } & \multicolumn{2}{c}{Polyvore-S} & \multicolumn{2}{c}{Polyvore-D} \\
\hline
\multicolumn{1}{c}{ } & \multicolumn{1}{c}{Train} & \multicolumn{1}{c}{Test} & \multicolumn{1}{c}{Train} & \multicolumn{1}{c}{Test}\\
\hline
\textbf{\#Nodes} & 204679 & 47854 & 71967 & 70035\\
\textbf{\#Edges} & 685024 & 129496 & 192545 & 165313\\
\textbf{Avg. Degree} & 6.693 & 5.412 & 5.35 & 4.721\\
\textbf{Median Degree} & 5 & 5 & 5 & 4\\
\textbf{Conn. Components} & 9226 & 4994 & 4837 & 11224\\
\textbf{Transitivity} & 0.328 & 0.732 & 0.654 & 0.806\\
\textbf{Avg. Cluster Coeff.} & 0.88 & 0.945 & 0.916 & 0.955\\
\hline
\end{tabular}
\end{table}

These numbers do not allow operating on the entire IRG at once, as it will saturate the GPU memory.
To overcome this issue, we first split the graph into clusters and only retain relevant partitions. Let $G_{\mathcal{O}}$ be the dataset ORG, as in Definition \ref{def:org}. Based on the idea of Cluster-GCN \cite{chiang2019clustergcn}, this ORG is partitioned, using the METIS clustering algorithm \cite{karypis1998fast}, in $\mathcal{P}$ subgraphs such that each of them contains about $\phi$ nodes (i.e.~outfits). We refer to the p-th partition of the ORG as $G_{\mathcal{O},p}$.
In our experiments we use $\phi = 50$.

\subsection{Training Setup}\label{training-setup}
Given an outfit $o=\{g_{1}, \cdots, g_{n}\}$, a training example is a triplet $(\hat{o},G_{\mathcal{G},o},\mathcal{C}_{\hat{o}})$ where $\hat{o}$ is a random permutation of the original outfit $o$. This means that the model sees a different shuffled version of the same outfit at every epoch, acting as a form of data augmentation and making the model invariant to garment ordering. To build the partitioned IRG $G_{\mathcal{G},o}$, relative to outfit $o$, we adopt the following steps:

\begin{enumerate}
\item 
  Find the partition $G_{\mathcal{O},p}=(V_p,\mathcal{E}_p)$ such that $o \in V_p$.
\item
  From $G_{\mathcal{O},p}$, build its IRG induction $\overline{G_{\mathcal{O},p}}=(\overline{V_p},\overline{\mathcal{E}_p})=G_{\mathcal{G},p}$ as in Definition \ref{def:org-to-irg}.
\item
  Remove all the edges $e_{i,j} \in \overline{\mathcal{E}_p}$ such that $g_i, g_j \in o, g_i \ne g_j$, i.e. all edges connecting elements of $o$. 
\end{enumerate}

Finally, for each garment $g_i$ in the outfit that has to be predicted, we build a candidate set $C_i$ containing $(n_C + n_R + 1)$ garments:


\begin{itemize}
\item
  $n_C$ random negatives $g_j \in G_{\mathcal{G},o}$ such that $g_j \notin o$ and $C(g_j) = C(g_i)$, i.e. belonging to the same category of the ground truth item.
\item
  $n_R$ random distractors $g_j \in G_{\mathcal{G},o}$ such that $g_j \notin o$.
\item
  $1$ garment representing the ground truth item $\Gamma_i$.
\end{itemize}

We refer to $\mathcal{C}_{\hat{o}} = \{\mathcal{C}_1, \cdots, \mathcal{C}_n\}$ as the set of candidate sets for each step.
In our experiments we use, unless expressly stated otherwise, $n_C=3$ and $n_R=5$.

The model is trained to output the sequence of garments in the outfit, starting from a garment seed and optimizing the loss defined in Eq. \ref{eq:gtn-obj} at each generation step.

 \begin{figure}[t]
	{\centering \includegraphics[width=\linewidth]{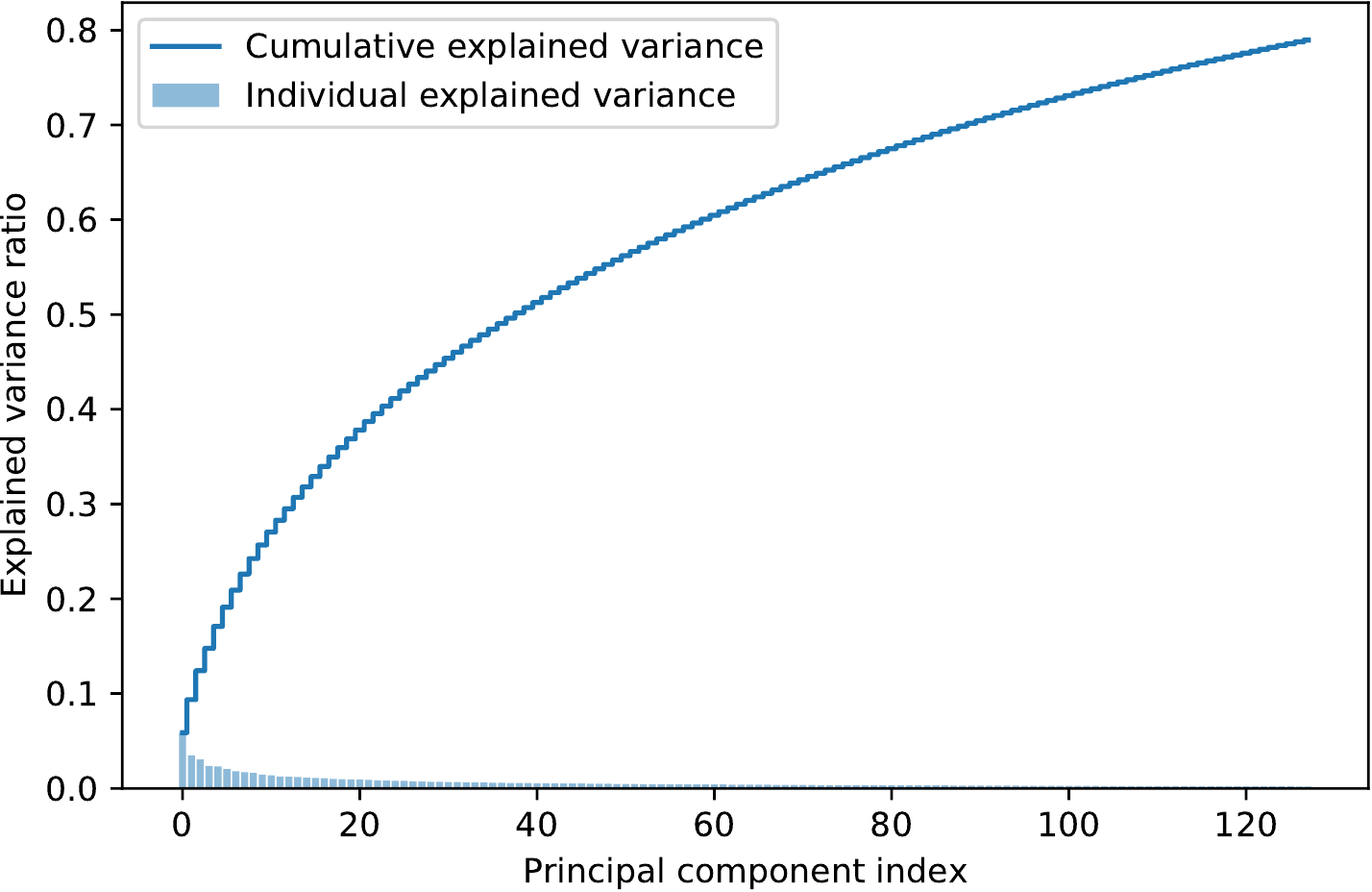} 
	}
	\caption{Explained variance for PCA computed on ResNet50 features extracted from Polyvore.}\label{fig:variance}
\end{figure}

\subsection{Training Details and Hyperparameters}\label{hyper-val}

If not otherwise stated, the following hyperparameter values are used throughout all the experiments.
We set the number of modules in the encoder block, i.e.~the aggregation radius, equal to $K_{enc}=2$. As for the decoder, we use a number of modules $K_{dec}=4$.
Garment images are projected using PCA to an embedding of dimension $d_e=128$. The explained variance for the PCA is depicted in Fig. \ref{fig:variance}. Such embeddings are projected by the transition layer into a hidden space with $d_m=256$ dimensions.
For multi-head attention, we use $H=8$ heads. 



The network is optimized with the Adam \cite{kingma2017adam} optimizer with learning rate $5 \cdot 10^{-4}$ and weight decay $5 \cdot 10^{-5}$. The learning rate is reduced by a $0.1$ factor each time the validation loss reaches a plateau. Dropout is applied to each dense layer and with value $0.35$. The model is trained for a maximum of $1000$ epochs with an early stopping strategy on the validation loss with a 10 epoch patience.

The GPU used during training is a Nvidia Titan RTX with 24GB memory. Half-precision was employed to lower memory requirements and training time.

 \begin{figure}[t]
 {\centering \includegraphics[width=\linewidth]{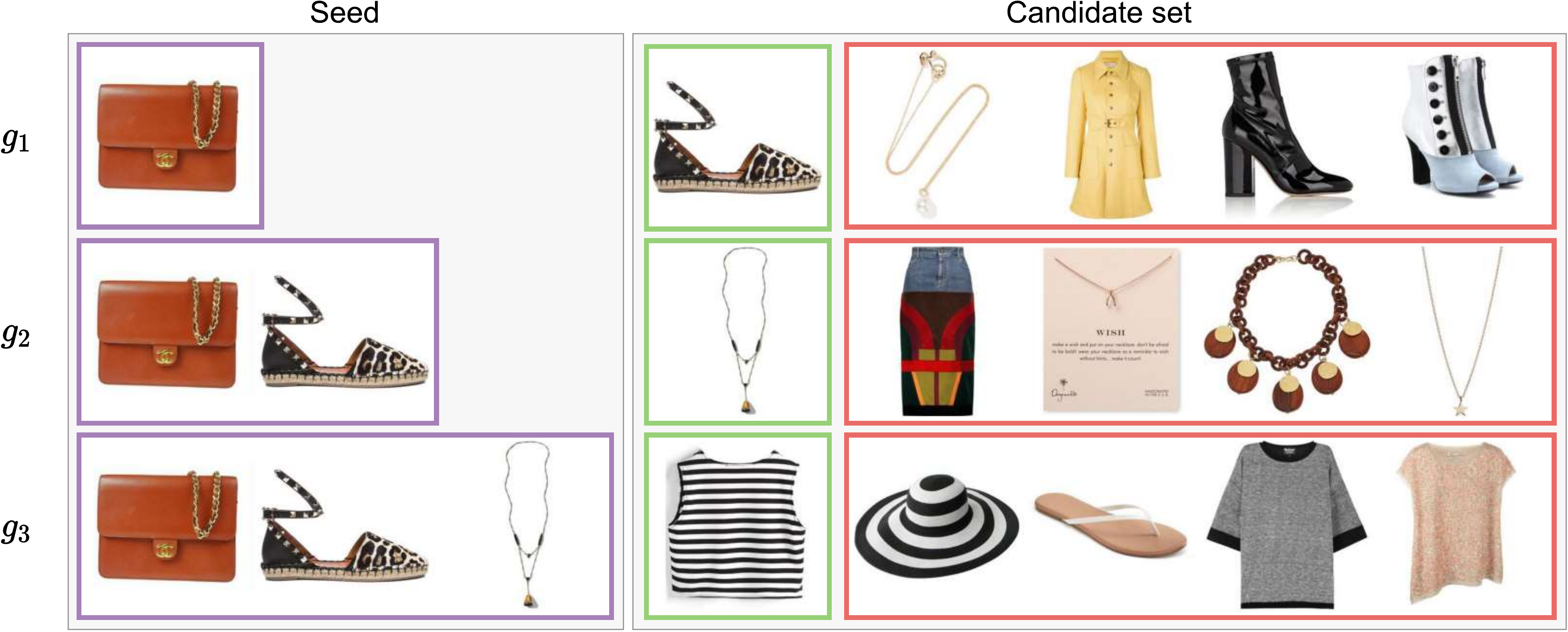} 
 }
 \caption{SIP is a task in which a model has to predict, for each item in a given outfit, the next one. Here is an example for the first 3 items of an outfit, along with their candidate sets from which to choose the next correct item. This task objective is to find the correct garment \(\Gamma_i=g_{i+1}\) given the seed \((g_1,\cdots,g_i)\).}\label{fig:sip-task}
 \end{figure}

\section{Evaluation tasks overview}\label{evaluation-tasks-overview}

The performance of TGNN is evaluated on three different tasks.
Our main focus is on outfit generation, for which we formalize an evaluation protocol named Seeded Item Prediction.
However, we also apply our model to standard compatibility tasks, namely Fill-In-The-Blank and Compatibility Prediction, which are widely adopted in literature \cite{vasileva2018learning}, \cite{han2017learning}, \cite{revanur2021semisupervised}. In the following we provide a brief overview of such tasks and relative evaluation metrics.

\textbf{Seeded Item Prediction (SIP)} is a task where the model is presented with a partial outfit and it has to incrementally predict the next complementary item given the previous ones from a collection of possible candidates (Fig. \ref{fig:sip-task}). The performance is evaluated using the accuracy in choosing the correct item.

\begin{figure}[t]
	{\centering \includegraphics[width=\linewidth]{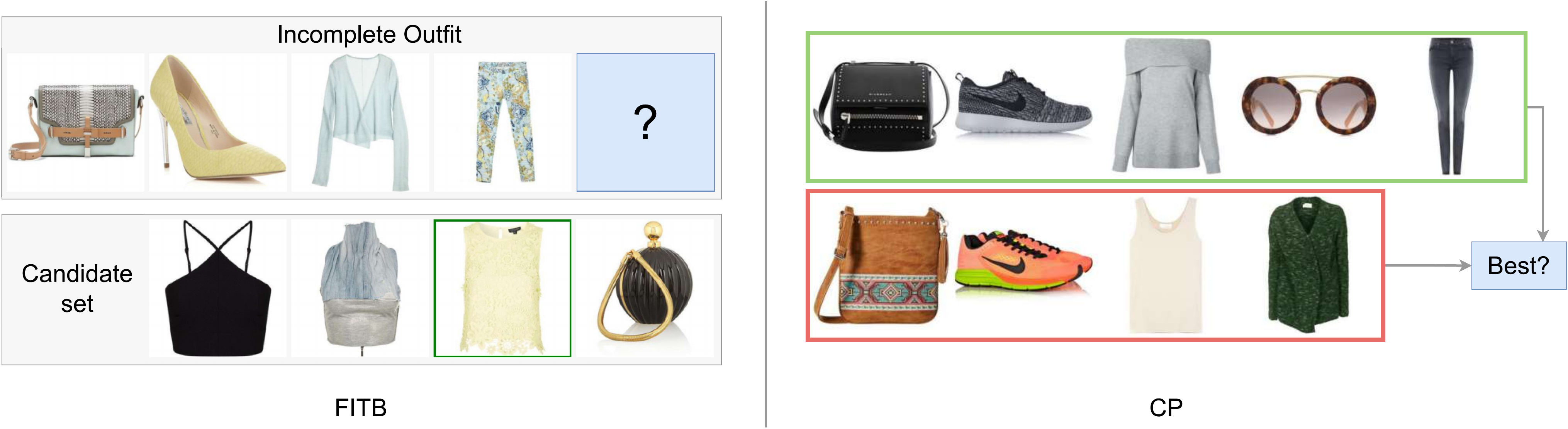} 
	}
	\caption{Visual illustration of the FITB (left) and the CP (right) tasks. \textbf{Left} -- In FITB, the task has to choose an item from a candidate set that is most compatible with a given incomplete outfit; the performance is evaluated by the accuracy of choosing the correct item, here visualized as the green one. \textbf{Right} -- In CP, the task scores how much a candidate outfit is compatible; to do this, it compares pairs of compatible (green) and incompatible (red) outfits, trying to assign a higher score to the compatible one.}\label{fig:fitb-cp-tasks}
\end{figure}

\textbf{Fill-In-The-Blank (FITB)} is a task in which the model is presented with an incomplete outfit missing one item. Along with the outfit, four candidate items (three incorrect and one correct), are provided as possible answers. The task is then to choose the item between the candidates that is the most compatible with the given incomplete outfit (Fig. \ref{fig:fitb-cp-tasks}). The results are evaluated as overall accuracy.
It must be noted that FITB is a special case of the SIP task: the two tasks are the same when the seed of SIP is the entire outfit minus one item and the possible answer set is composed by four items. In this sense, the commonly adopted FITB task is an oversimplification of SIP.

\textbf{Compatibility Prediction (CP)} is a task where the model has to predict whether a candidate outfit is compatible or not (Fig. \ref{fig:fitb-cp-tasks}). Outfits are scored to assess whether their constituting items are compatible with each other.
The task is performed by feeding to the model two different outfits, namely a positive ground truth outfit, which is known to be compatible, and a negative random outfit. The model is requested to score the two outfits and the task is considered successful when the positive one is scored higher than the negative one.
The performance is evaluated using the area under the receiver operating characteristic curve (AUROC).

\subsection{Test Setup}\label{test-setup}

Since the TGNN architecture works with triplets containing an outfit, a graph partition and a candidate set in the form of \((o,G_{\mathcal{G},o},\mathcal{C}_o)\), like the ones described in Section \ref{training-setup}, these must be carefully constructed for each evaluation task.

For the Fill-In-The-Blank task, given the original complete outfit $o=\{g_{1},\cdots,g_b,\cdots,g_{n}\}$, where $g_b$ is the blank garment placeholder corresponding to the item to be predicted, the triplet will be $(\hat{o},G_{\mathcal{G},o},\mathcal{C}_{\hat{o}})$ where:

\begin{itemize}
\item
  $\hat{o}$ is the incomplete outfit, that is, $o$ without the blank item $g_b$.
\item
  $G_{\mathcal{G},o}$ is a graph partition built in the same way as described in Section \ref{training-setup}. However, instead of identifying the partition that exactly contains $o$, we retrieve the most similar garments in the training set and take the partition containing the maximum number of such garments. This is done because, in general, a test outfit might not be present in the training set.
  \review{By doing so, we assume that test garments have some sufficiently similar items in the training graph, which we believe to be a reasonable assumption. Other methods from the state of the art not depending on a garment graph may not make this assumption.}
  \item $\mathcal{C}_{\hat{o}} = \mathcal{C}_b$ containing the four possible answers, among which there is the correct garment. Note that there is a single candidate set, since in the FITB task we only need to perform one prediction step to complete the outfit.
\end{itemize}

For the Compatibility Prediction task, instead, we need to identify a graph partition for both the positive and negative outfits. Both partitions are taken looking for the most similar items in the training set, as in the FITB configuration.

We then generate a random partition of the garments and ask the model to recreate the outfit from a single-element seed. We average the likelihood of the correct garment at each generation step to obtain the final compatibility prediction score.
In order to do so, we need to define a candidate set $\mathcal{C}_i$ for each step. Each one is composed of the ground truth item $\Gamma_i$ and three other random garments belonging to $G_{\mathcal{G}, o}$.


Finally, for the Seeded Item Prediction task, the setup is the same as Section \ref{training-setup} without the initial random permutation. Also in this case, the graph partition is obtained by taking the one with the maximum number of most similar items to the partial outfit in the training set.


\subsection{Baselines}

To validate the hypothesis that contextual information, i.e.~the Item Relation Graph, and self-attention play an important role for compatibility learning, TGNN is compared with previously reported approaches on the Polyvore dataset:

\begin{itemize}
\item
  \textbf{Siamese Network} - The approach described by \emph{Veit et al.}~\cite{veit2015learning} that estimates pairwise compatibility based on co-occurrence in large-scale user behavior data.
\item
  \textbf{Bi-LSTM} - \emph{Han et al.} \cite{han2017learning} was the first work to consider an outfit as a whole, representing it as an ordered sequence. Taking multi-modal data as input, it can calculate the compatibility scores of outfits by iteratively predicting the next item.
\item
  \textbf{CSN T1:1} - Learns a pairwise category-dependent transformation using the approach of \emph{Veit et al.} \cite{veit2017conditional} to project a general embedding to a single type-specific space, measuring compatibility between two item categories.
\item
  \textbf{Type Aware} - This method, described in \emph{Vasileva et al.} \cite{vasileva2018learning}, focuses on learning item image embeddings that respect the item type by measuring pairwise item similarity in different semantic subspaces.
\item
  \textbf{SCE-Net} - Introduced in \emph{Tan et al.} \cite{tan2019learning}, it jointly learns pairwise different representation subspaces for different similarity conditions without explicit supervision.
\item
  \textbf{CSA-Net} - Approach introduced by \emph{Lin et al.} \cite{lin2020fashion} that models outfit compatibility by encoding category pairs information to each item embedding vector.
\item
  \textbf{Pseudo-Label} - Described in \emph{Revanur et al.} \cite{revanur2021semisupervised}, this method, developed for semi-supervised settings, learns item embeddings by generating pseudo positive and negative pairs on-the-fly during each training epoch. These pairs can be based on different characteristics, like colors, shapes, etc. while discarding the item category notion.
  
\item
  \textbf{ADDE-O} - The method \cite{hou2021learning} is based on learning disentangled features for specific attribute classes. This allows to perform targeted manipulations exploiting a memory bank of attribute prototypes.

\item
  \textbf{MUFIN} - Recently proposed in \cite{mittal2022multi}, the method addresses the task of Extreme Classification, i.e. a training strategy involving millions of labels leveraging both visual and textual descriptors.
  
\item
  \textbf{OutfitTransformer} - This paper \cite{sarkar2022outfittransformer} proposes to obtain outfit-level representations with transformers to address the tasks of compatibility prediction and fill in the blank, as well as a complementary outfit retrieval taks.
  
\end{itemize}

We also report variants of some of the methods, referred to as \textit{VSE}, which add a visual-semantic embedding, as described in \cite{vasileva2018learning}, jointly learned with the compatibility embedding.

\begin{table}[t]
\caption{Accuracy on the Seeded Item Prediction task for both Polyvore-D and Polyvore-S datasets.}
\label{tab:sip-results}
\centering
\begin{tabular}[t]{l|cc}
\hline
 & Polyvore-D & Polyvore-S \\
\hline
Random & 11.0 & 11.2 \\
Type Aware \cite{vasileva2018learning} & 43.3 & 42.0 \\
TGNN & 51.3 & 55.6 \\
\hline
\end{tabular}
\end{table}

\begin{table*}[t]
	\caption{\label{tab:result-table}Comparison of TGNN against state-of-the-art methods. Some of these rely on additional supervision such as text embeddings and explicit category supervision. FITB accuracy and Compatibility AUROC are reported. \emph{Higher is better}.}
	\centering
		\begin{tabular}[t]{l|c|ccccc}
   \hline
			\multicolumn{3}{c}{ } & \multicolumn{2}{c}{Polyvore-D} & \multicolumn{2}{c}{Polyvore-S} \\
   \hline
			Method & Text labels & Explicit category & FITB acc & CP AUROC & FITB acc & CP AUROC\\
   \hline
			Siamese Network \cite{vasileva2018learning} & $\times$ & $\times$ & 51.8 & 0.81 & 52.9 & 0.81\\
			Bi-LSTM + VSE \cite{han2017learning} & $\checkmark$ & $\times$ & 39.4 & 0.62 & 39.7 & 0.65\\
			CSN T1:1 \cite{veit2017conditional} & $\times$ & $\checkmark$ & 52.5 & 0.82 & 54.0 & 0.83\\
			CSN T1:1 + VSE \cite{veit2017conditional} & $\checkmark$ & $\checkmark$ & 53.0 & 0.82 & 54.5 & 0.84\\
			Type Aware \cite{vasileva2018learning} & $\checkmark$ & $\checkmark$ & 55.65 & 0.84 & 58.83 & 0.86\\
			SCE-Net (avg) \cite{lin2020fashion} & $\checkmark$ & $\times$ & 53.67 & 0.82 & 59.07 & 0.88\\
			Pseudo-Label \cite{revanur2021semisupervised} & $\times$ & $\times$ & 54.6 & 0.84 & 57.9 & 0.89\\
			CSA-Net \cite{lin2020fashion} & $\times$ & $\checkmark$ & 59.26 & 0.87 & 63.73 & 0.91\\
			ADDE-O \cite{hou2021learning} & $\times$ & $\times$ & 60.53 & 0.88 & 65.16 & 0.93 \\
			MUFIN \cite{mittal2022multi} & $\checkmark$ & $\times$ & 64.17 & - & - & -\\
			OutfitTransformer \cite{sarkar2022outfittransformer} & $\checkmark$ & $\times$ & 59.48 & 0.88 & 67.10 & 0.93\\ \hline
			\textbf{TGNN-18} & $\times$ & $\times$ & 65.36 & 0.93 & 68.72 & 0.95\\
			\textbf{TGNN} & $\times$ & $\times$ & \textbf{65.74} & \textbf{0.94} & \textbf{69.03} & \textbf{0.96}\\
   \hline
	\end{tabular}
\end{table*}

\begin{figure}[t]
	{\centering
 \includegraphics[height=0.18\linewidth]{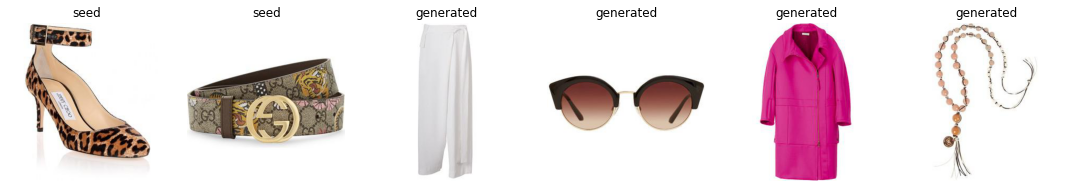} \\
 \includegraphics[height=0.18\linewidth]{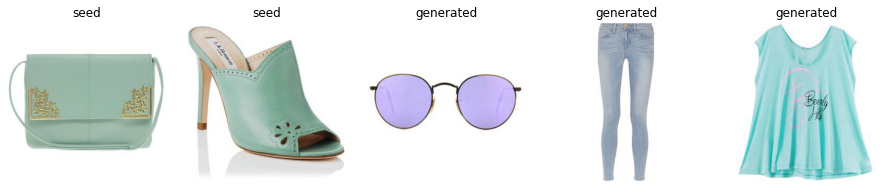}  \\
 \includegraphics[height=0.18\linewidth]{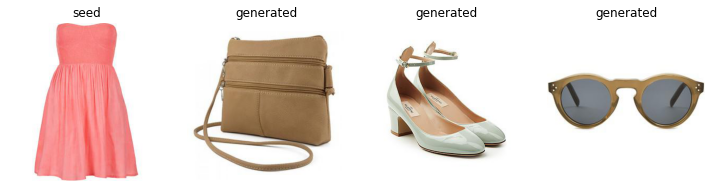} \\
 \includegraphics[height=0.18\linewidth]{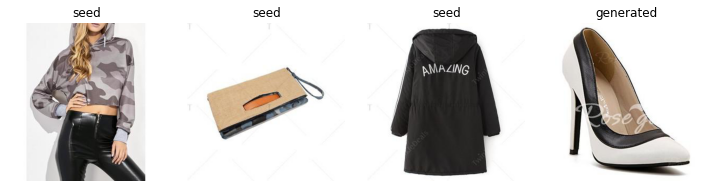} \\
	}
	\caption{Qualitative results obtained by TGNN on Polyvore-D. Images marked as \textit{seed} are garments belonging to the same outfit, whereas \textit{generated} ones are items suggested by TGNN to complete the outfit.}\label{fig:qualitative}
\end{figure}

\begin{figure}[t]
	{\centering
 \includegraphics[height=0.18\linewidth]{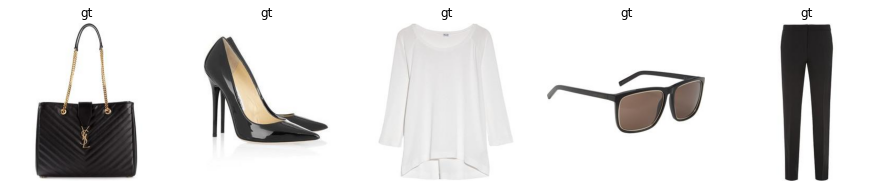} \\
 \includegraphics[height=0.18\linewidth]{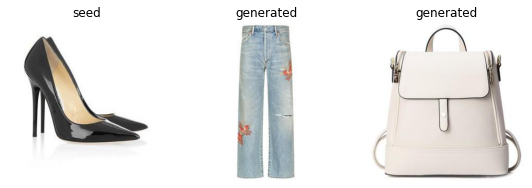} \\
 ~~ \\
 \includegraphics[height=0.18\linewidth]{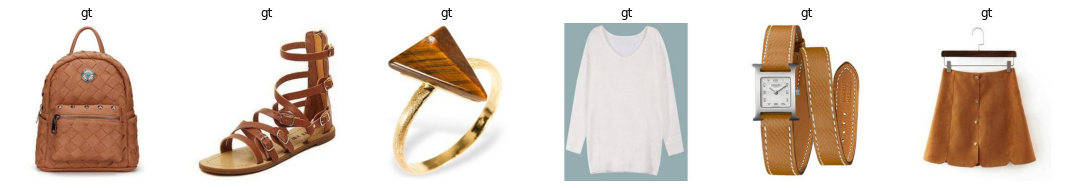} \\
 \includegraphics[height=0.18\linewidth]{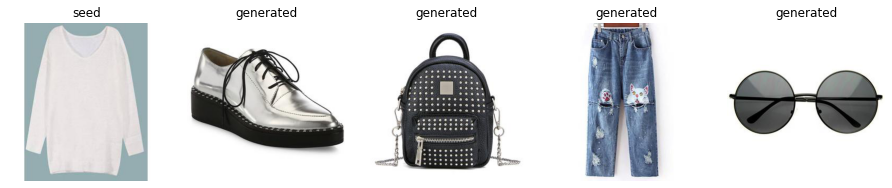} \\
 }
	\caption{Failure cases. In the first example, the model has failed to complete the outfit, picking only a bottom and a bag without choosing a top. In the second one instead TGNN has built an outfit with a different style than one ground truth one.}\label{fig:failure}
\end{figure}

\section{Results}
\label{results}
We first evaluate our model on the newly introduced task of Seeded Item Prediction (SIP). The task was created to evaluate the outfit generation capabilities of TGNN. As illustrated in Fig.~\ref{fig:sip-task}, given an outfit $o=\{g_1,\cdots,g_n\}$, the task objective is to correctly predict the next item $\Gamma_i=g_{i+1}$ from a candidate set, given the previous items $\{g_1,\cdots,g_i\}$. Tab.~\ref{tab:sip-results} shows how TGNN performs on this task, along with two baselines, by measuring the accuracy of choosing the correct item.
Since SIP is a novel evaluation task proposed in this work, there are no available results from previous works. To provide a solid baseline, the task was evaluated with the publicly released code of the Type Aware architecture\footnote{Original code available at \url{https://github.com/mvasil/fashion-compatibility}}.
To generate the results, we trained the model with the same hyperparameter values as described in the original paper~\cite{vasileva2018learning}. We provide also a random baseline as a lower bound.

TGNN performs well on this difficult task, showing that it is capable of correctly generating outfits. There is a gap between the accuracy obtained with the two Polyvore versions. We attribute this to the attention mechanism, whose characteristic is to learn how different items are related among themselves. Therefore, the presence, or absence, in the test set of items seen during the training phase could most likely influence this mechanism, and, in turn, the whole model performance.

In Fig.~\mbox{\ref{fig:qualitative}} we report qualitative results of correctly generated outfits by TGNN in the Seeded Item Prediction setting. It can be seen that TGNN is able to generate outfits with a variable number of outfits, starting from a variable number of seeds.
We also show a few failure cases in Fig.~\mbox{\ref{fig:failure}}. In the first failure case, the model emits the stop sign before completing the outfit, lacking a top garment to complete the outfit. In the second example, the model generates an outfit with a style different from the ground truth one. This behaviour is easily avoidable by adding more seed items to the input.

We then evaluate TGNN for the tasks of Fill-In-The-Blank (FITB) and Compatibility Prediction (CP). As shown in Tab.~\ref{tab:result-table}, TGNN outperforms previous state of the art methods on both tasks.
These results suggest that the graph structure provides valuable information for this kind of task. In particular for FITB it is indeed to be expected that, within an IRG, the missing garment (or one similar enough) will be found close to its complementary items.
At the same time, for Compatibility Prediction we believe that the multi-head self-attention mechanisms of TGNN plays an important role: by learning to model compatibilities between an item and all the other ones within the same outfit, the model implicitly learns a global outfit compatibility notion by combining all of these item compatibilities.

Since several methods adopt a ResNet18 backbone \mbox{\cite{he2016deep}} rather than a ResNet50 like ours, we also show results for a variant of TGNN using ResNet18 features to establish a fair comparison. Apart from the backbone the whole training and evaluation procedure is left unchanged. Indeed, using ResNet18 features there is a slight performance drop but TGNN still manages to achieve state of the art results.


\begin{table}[t]
\caption{\label{tab:ablation-phi}TGNN results for Seeded Item Prediction and Compatibility Prediction, calculated for different \(\phi\) values.
}
\centering
\begin{tabular}[t]{l|rr|rr}
\hline
\multicolumn{1}{c}{ } & \multicolumn{2}{c}{Polyvore-D} & \multicolumn{2}{c}{Polyvore-S} \\
\hline
 & SIP & CP & SIP & CP\\
\hline
$\phi=25$ & 45.7 & 0.88 & 52.4 & 0.89 \\
$\phi=50$ & \textbf{51.3} & \textbf{0.94} & \textbf{55.6} & \textbf{0.96} \\
$\phi=100$ & 44.0 & 0.78 & 49.7 & 0.80 \\
$\phi=200$ & 34.8 & 0.78 & 42.9 & 0.80 \\
\hline
\end{tabular}
\end{table}

\subsection{Ablation Studies}\label{ablation-studies}

To better analyze the TGNN behavior, it is useful to correlate the performance with those hyperparameters peculiar to the proposed architecture, namely $\phi$ -- the one controlling the partition sizes -- and $K_{enc}$ -- the one defining how far from a node the information gets aggregated into the said node, i.e.~the aggregation radius.

\subsubsection{Partition Size}
\label{partition-size-study}
In Tab.~\ref{tab:ablation-phi}, the TGNN results for different $\phi$ values are shown.
It is clear that, when increasing $\phi$, i.e.~when the partitions $G_{\mathcal{G},o}$ grow larger, TGNN performance decreases. The main reasons are:

\begin{enumerate}
\item
  The encoder block is no longer able to learn, and thus to integrate into the nodes useful relational information from the surrounding context. In other words, there is too much information and the encoder block fails to handle it.
\item
  The \emph{Encoder-Decoder Attention} of each decoder module relates each item $g_i \in o$ with all the nodes of $G_{\mathcal{G},o}$; with higher $\phi$ values, the number of nodes grows exponentially, thus, making the Encoder-Decoder Attention layer increasingly complex.
\end{enumerate}

On the contrary, when lowering the partition size too much, the accuracy decreases since not enough context is given to the model.


\begin{table}
\caption{\label{tab:ablation-k-enc}TGNN performance on Polyvore-D for different \(K_{enc}\) values.}
\centering
\begin{tabular}[t]{l|rr|rr|r}
\hline
\multicolumn{1}{c}{ } & \multicolumn{2}{c}{Polyvore-D} & \multicolumn{2}{c}{Polyvore-S} \\
\hline
 & SIP & CP & SIP & CP & Train time\\
\hline
$K_{enc}=2$ & 35.9 & 0.77 & 32.6 & 0.83 & 3h 27m \\
$K_{enc}=3$ & 42.5 & 0.86 & 39.0 & 0.91 & 5h 50m \\
$K_{enc}=4$ & \textbf{51.3} & 0.94 & \textbf{55.6} & \textbf{0.96} & 10h 22m \\
$K_{enc}=5$ & \textbf{51.3} & \textbf{0.95} & 55.5 & 0.94 & 17h 31m \\
\hline
\end{tabular}
\end{table}

\subsubsection{Aggregation Radius}
\label{aggregation-radius-study}
In Tab.~\ref{tab:ablation-k-enc}, the results obtained by TGNN are shown, at increasing values of the aggregation radius $K_{enc}$. The experiments were carried out with all the hyperparameter values fixed to the ones listed in Sec.~\ref{hyper-val}.
It is clear that increasing $K_{enc}$ yields better results.
In fact, $K_{enc}$ controls how many encoder modules are used in the encoder block, which, in turn, affects how far from a given node the information gets aggregated, i.e.~from the $K_{enc}$-neighborhood of the said node.
Increasing $K_{enc}$ does indeed help the encoder block at managing the information in the graph, while learning useful relationships between loosely coupled outfits.
Eventually, the benefits of increasing $K_{enc}$ saturate, bringing negligible or no benefits, yet requiring a much longer training time (Tab.~\ref{tab:ablation-k-enc}).

\section{Conclusions}\label{conclusions}

This work focused on the problem of outfit generation starting from a garment seed, which consists in proposing a set of garments that, combined with the seed, assembles a compatible outfit. To this aim, \emph{Transformer-based Graph Neural Network} is proposed as a novel architecture which combines the capabilities of Graph Neural Networks with the powerful Transformer model.
The proposed method is capable of learning the complex relations existing between items and outfits by employing graph representations and attention mechanisms.
Through message propagation and self-attention, an item compatibility notion is refined by aggregating the interaction information derived from its neighbors. This knowledge is then used in the generation phase to iteratively choose the item that is most compatible with the seed and the previously chosen ones.
Extensive experiments on the Polyvore dataset have demonstrated the rationality and the effectiveness of TGNN, which outperforms previous models in outfit compatibility tasks.

\section{Further developments}\label{further-developments}

Since the iterative generation mechanism is inspired to the one employed in many NLP models, we plan to improve the generation performance with search strategies inspired by the field of NLP, such as Beam Search.
Another matter worth investigating is making the architecture work with a dynamic IRG. In other words a graph that can change over time by adding new nodes or edges. Being able to handle such real-world situations could prove fundamental to evolve a research model in a production ready one.
Furthermore, to resolve the Encoder-Decoder Attention issue when the $\phi$ value increases, further studies could focus on creating masks in order to intelligently limit the number of partition nodes to which any outfit item is related to.
Finally, an assessment from domain experts as an additional term of comparison between methods could help understanding which aspects are taken into account while composing the outfit, such as style, color, etc.


\ifCLASSOPTIONcompsoc
  \section*{Acknowledgments}
\else
  \section*{Acknowledgment}
\fi

This work was partially supported by the Italian MIUR within PRIN 2017, Project Grant 20172BH297: I-MALL - improving the customer experience in stores by intelligent computer vision.

\bibliographystyle{IEEEtran}
\bibliography{sample-base}

\begin{IEEEbiography}[{\includegraphics[width=1in,height=1.25in,clip,keepaspectratio]{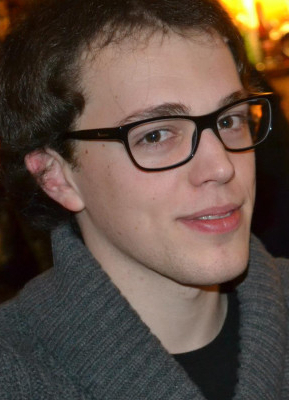}}]	{Federico Becattini} obtained his PhD in 2018 from the University of Florence under the supervision of Prof. Alberto Del Bimbo and Prof. Lorenzo Seidenari. Currently he is a PostDoc at the University of Florence, where he is involved in numerous collaborations, mostly focusing on Autonomous Driving, Scene Understanding and Fashion Recommendation. He attended several international conferences both as speaker and volunteer, as well as summer schools. He served to the scientific community as a reviewer for scientific journals and conferences. he is Associate Editor of the International Journal of Multimedia Information Retrieval.
\end{IEEEbiography}

\begin{IEEEbiography}[{\includegraphics[width=1in,height=1.25in,clip,keepaspectratio]{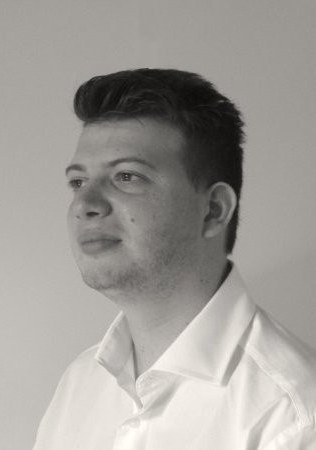}}]{Federico Maria Teotini} received a master degree cum laude in 2022 in computer engineering from the University of Florence with the thesis “Graph-Transformer Network for Outfit Generation”.
\end{IEEEbiography}

\begin{IEEEbiography}[{\includegraphics[width=1in,height=1.25in,clip,keepaspectratio]{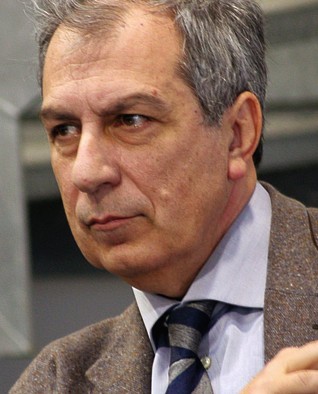}}]{Alberto Del Bimbo} is a Full Professor of Computer Engineering, and the Director of the Media Integration and Communication Center with the University of Florence. His scientific interests are multimedia information retrieval, pattern recognition, and computer vision. From 1996 to 2000, he was the President of the IAPR Italian Chapter and the Member-at-Large of the IEEE Publication Board from 1998 to 2000. He was the General Co-Chair of ACMMM2010 and ECCV2012. He was nominated ACM Distinguished Scientist in 2016. He received the SIGMM Technical Achievement Award for Outstanding Technical Contributions to Multimedia Computing, Communications and Applications. He is an IAPR Fellow, and Associate Editor of Multimedia Tools and Applications, Pattern Analysis and Applications, Journal of Visual Languages and Computing, and International Journal of Image and Video Processing, and was Associate Editor of Pattern Recognition, IEEE Transactions on Multimedia, and IEEE Transactions on Pattern Analysis and Machine Intelligence.
\end{IEEEbiography}


\end{document}